%% file: sample.tex
\documentclass[twoside,11pt]{article}

\usepackage{blindtext}

\usepackage[preprint]{jmlr2e}
\usepackage[printonlyused]{acronym}
\usepackage{caption}
\input{acronyms}

\newcommand{\frameworkName}{\textsc{RSL-RL}\xspace}

\usepackage[nameinlink]{cleveref}
\crefname{equation}{Eq.}{Eqs.}
\crefname{figure}{Fig.}{Figs.}
\crefname{section}{Sec.}{Sec.}
\crefname{appendix}{App.}{App.}
\crefname{table}{Tab.}{Tabs.}
\crefname{algorithm}{Algo}{Algo}
\crefname{thm}{Thm}{Thm}
\Crefname{thm}{Thm}{Thm}
\crefname{prop}{Prop}{Prop}

\usepackage{lastpage}
\jmlrheading{23}{2025}{1-\pageref{LastPage}}{8/25; Revised 1/26}{2/26}{25-0000}{Clemens Schwarke, Mayank Mittal, Nikita Rudin, David Hoeller, and Marco Hutter}

\ShortHeadings{\frameworkName: A Learning Library for Robotics Research}{Schwarke, Mittal, Rudin, Hoeller, and Hutter}
\firstpageno{1}

\begin{document}

\title{\frameworkName: A Learning Library for Robotics Research}

\author{\name Clemens Schwarke$^{1, 2}$ \email cschwarke@ethz.ch \\
        % \addr Department of Mechanical and Process Engineering (D-MAVT) \\
        % ETH Z\"{u}rich, 8092 Z\"{u}rich, Switzerland
        % \AND
        \name Mayank Mittal$^{1, 2}$ \email mittalma@ethz.ch \\
        % \addr Department of Mechanical and Process Engineering (D-MAVT) \\
        % ETH Z\"{u}rich, 8092 Z\"{u}rich, Switzerland
        % \AND
        \name Nikita Rudin$^{1, 2, 3}$ \email nikita@flexion.ai \\
        % \addr Flexion Robotics, 8050 Z\"{u}rich, Switzerland
        % \AND
        \name David Hoeller$^{1, 2, 3}$ \email david@flexion.ai \\
        % \addr Flexion Robotics, 8050 Z\"{u}rich, Switzerland
        % \AND
        \name Marco Hutter$^{1}$ \email mahutter@ethz.ch \\
        \addr $^{1}$ETH Z\"{u}rich, $^{2}$NVIDIA, $^{3}$Flexion Robotics
        % $^\dagger$ Project Lead, $^*$ Core Developer, $^\ddagger$ Project Mentor
        % \addr Department of Mechanical and Process Engineering (D-MAVT) \\
        % ETH Z\"{u}rich, 8092 Z\"{u}rich, Switzerland
       }

\editor{My editor}

\maketitle

\begin{abstract}%   <- trailing '%' for backward compatibility of .sty file

\frameworkName is an open-source \acl{RL} library tailored to the specific needs of the robotics community. Unlike broad general-purpose frameworks, its design philosophy prioritizes a compact and easily modifiable codebase, allowing researchers to adapt and extend algorithms with minimal overhead.
The library focuses on algorithms most widely adopted in robotics, together with auxiliary techniques that address robotics-specific challenges.
Optimized for GPU-only training, \frameworkName achieves high-throughput performance in large-scale simulation environments.
Its effectiveness has been validated in both simulation benchmarks and in real-world robotic experiments, demonstrating its utility as a lightweight, extensible, and practical framework to develop learning-based robotic controllers.
The library is open-sourced at: \href{https://github.com/leggedrobotics/rsl_rl}{\texttt{\small{https://github.com/leggedrobotics/rsl\_rl}}}.
\end{abstract}


\begin{figure}[h]
    \centering
    \captionsetup{width=0.9\linewidth}
    \includegraphics[width=0.9\linewidth]{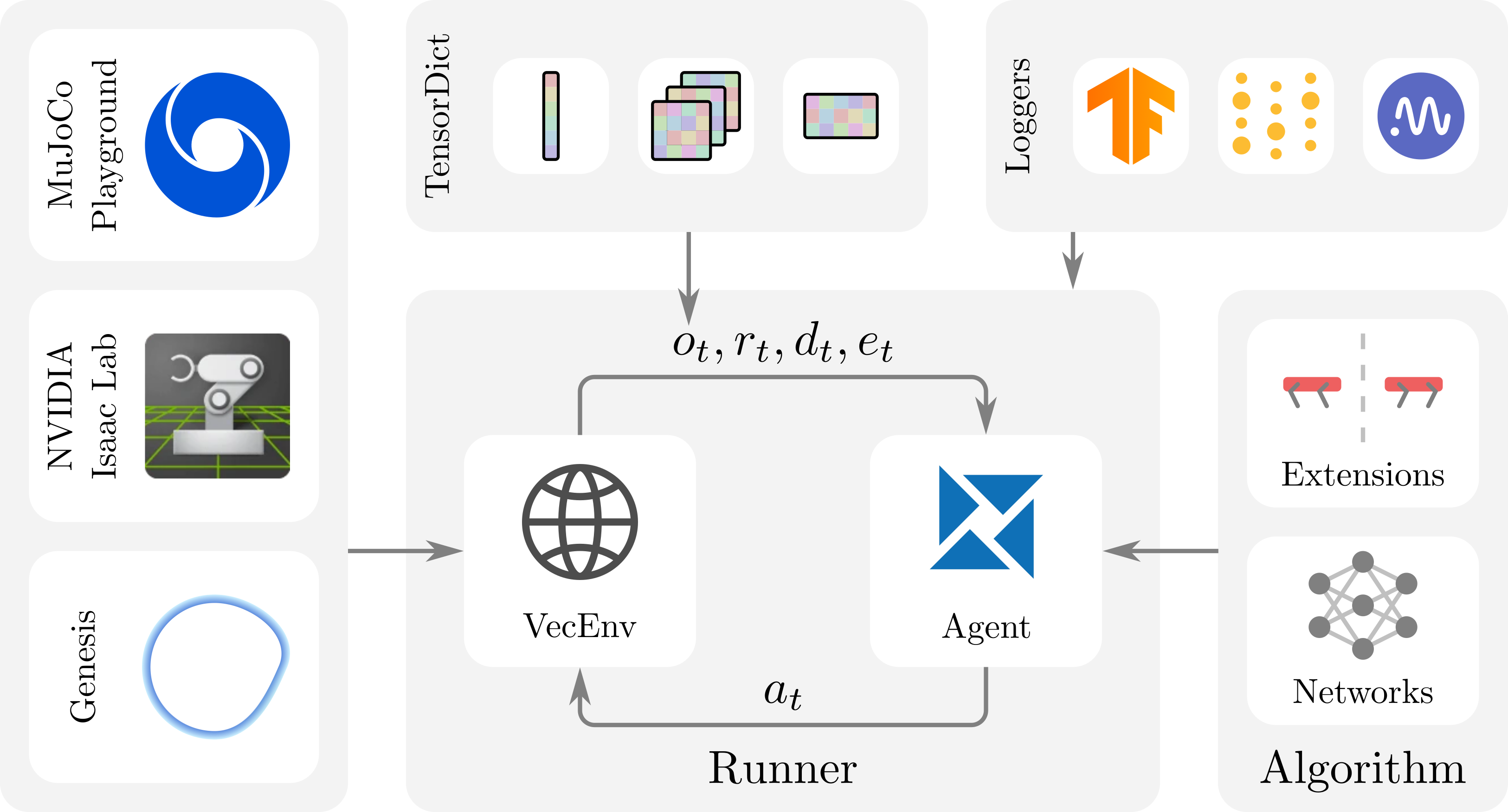}
    \caption{\textbf{Overview of the framework.} \frameworkName consists of three main components: \texttt{Runners}, \texttt{Algorithms}, and \texttt{Networks}, which can be easily modified independently. The framework comes with support for common logging choices and useful extensions for robotics.}
    \label{fig:main}
\end{figure}

\section{Introduction}

\ac{RL} has proven highly effective in tackling complex planning and control problems, often surpassing traditional model-based approaches~\citep{hwangbo2019learning, miki2022learning, handa2023dextreme}. These advances have been supported by both the development of better learning algorithms~\citep{schulman2017proximal} and the emergence of GPU-accelerated physics simulators~\citep{makoviychuk2isaac,mittal2023orbit,zakka2025mujoco} that enable large-scale parallelized training on consumer-grade hardware.

Although numerous \ac{RL} libraries are available~\citep{raffin2021stable, liang2018rllib, huang2022cleanrl, serrano2023skrl, rl-games2021, bou2023torchrl}, most are developed for the larger machine learning community. Their breadth of algorithms facilitates benchmarking, but the resulting modularity can make the code harder to adapt and extend. Consequently, adoption in robotics has been limited. Although robotics benefits from advances in fundamental algorithmic research, researchers often prioritize a compact and easily modifiable codebase that can efficiently benefit from large-scale simulation. At the same time, robotics applications frequently require specialized features, such as tools for distilling learned policies for real-world deployment.

\frameworkName addresses these limitations by deliberately adopting a minimalist yet powerful design, centered on a curated set of widely adopted state-of-the-art algorithms and features tailored for roboticists.
The core characteristics of the framework are as follows:
\begin{itemize}
    \item \textbf{Minimalist design:} A compact, readable codebase with clear extension points (runners, algorithms, networks) for rapid research prototyping.
    \item \textbf{Robotics-first methods:} \ac{PPO} and DAgger-style \ac{BC}, along with auxiliary techniques (symmetry augmentation and curiosity-driven exploration) that are often absent from general \ac{RL} libraries.
    \item \textbf{High-throughput training:} A GPU-only pipeline for large-scale batched training with native multi-GPU/multi-node support.
    \item \textbf{Proven and packaged:} Used in numerous robotics research publications, and integrated with various GPU-accelerated robotic simulation frameworks such as NVIDIA Isaac Lab, MuJoCo Playground, and Genesis for out-of-the-box use.
\end{itemize}

\section{Features}

In the following, we introduce the main functionalities of \frameworkName at the time of writing. While more algorithms and features are planned for the future, the library is meant to be adaptable, and our main priority is to keep it simple and easy to use.

\subsection{Algorithms}

Currently, \frameworkName includes two algorithms: \ac{PPO}~\citep{schulman2017proximal} and a \ac{BC} algorithm similar to DAgger~\citep{ross2011reduction}. \ac{PPO} is a model-free, on-policy \ac{RL} method that has become a standard in robot learning for its robustness and simplicity~\citep{shakya2023reinforcement}. It can be used to learn complex tasks from scratch, without requiring prior knowledge or demonstrations. In robotics, \ac{PPO} is most commonly applied to continuous control problems such as locomotion, manipulation, planning, and navigation.

The \ac{BC} algorithm is a supervised learning method used to distill the behavior of an expert policy into a student policy. It iteratively collects data by rolling out the student policy, relabels the data with expert actions, and trains the student on it. This algorithm is particularly useful after \ac{RL} training with \ac{PPO}, if the requirements for training and hardware deployment differ. In such cases, the behavior of an \ac{RL} agent relying on information only available in simulation can be distilled into a policy that does not rely on it.

\subsection{Auxiliary Techniques}

Alongside the main algorithms, the library includes techniques to improve performance in robotics applications. Currently, two such extensions are implemented, but this selection will be expanded in the future. The first technique, symmetry augmentation, accelerates sample generation by augmenting the collected data with mirrored states, exploiting the robot's physical symmetries~\citep{mittal2024symmetry}. This leads to more symmetric behaviors, an effect that can be reinforced with an additional symmetry loss.

A second technique improves exploration in sparse reward settings using a curiosity-driven intrinsic reward based on \ac{RND}~\citep{burda2019exploration}. However, in contrast to the original \ac{RND} formulation, the included implementation computes the reward using only a subset of the full state of the system. This modification allows the agent's curiosity to be focused on specific parts of the state space. Curiosity-driven exploration can substantially reduce the need for hand-engineered dense rewards. For instance, \citet{schwarke2023curiosity} trained a robot to open a door while balancing on two legs with a single binary task reward.

\subsection{Utilities}

\frameworkName provides several options for logging and evaluating experiments. The most straightforward option is TensorBoard~\citep{abadi2016tensorflow}, which facilitates local experiment logging. For a more sophisticated evaluation, the library supports Weights \& Biases~\citep{wandb} and Neptune~\citep{neptune}. These are excellent choices for training on compute clusters, as they allow experiments to be monitored live online. However, these services require a user account and involve uploading data to their respective cloud platforms. The library also supports distributed training over multiple nodes and multi-GPUs for both of its algorithms, enabling large-scale experimentation.

\section{Implementation Details}

The framework is organized into three main components, outlined in Fig.~\ref{fig:main}: (i) the \texttt{Runner} that manages environment stepping and agent learning, (ii) the \texttt{Algorithm} that defines the learning agent, and (iii) the \texttt{Network} architectures used by the algorithm. For most applications, users only need to modify these three files. Implemented in PyTorch~\citep{paszke2019pytorch} and Python, the framework is designed to be intuitive and easily extensible.

The framework defines its own \texttt{VecEnv} interface, which requires environments to implement the same-step reset mode~\citep{towers2024gymnasium}. The \texttt{step} method must return PyTorch tensors, and the observations must be structured as a \emph{TensorDict}~\citep{bou2023torchrl}. TensorDicts provide a flexible dictionary-like container for batched tensors. They enable selective routing of observation components to different network modules and naturally support auxiliary techniques such as \ac{RND} or latent reconstruction for regularization.

The included \ac{PPO} algorithm incorporates several implementation subtleties highlighted in~\cite{shengyi2022the37implementation}, such as the proper handling of episodic timeouts. In large-batch training, many environments tend to terminate simultaneously at the beginning of training, producing correlated rollouts. To mitigate this issue, the framework supports random early termination of episodes during initialization, which improves sample diversity and stabilizes learning. For both \ac{RL} and distillation settings, recurrent networks are supported through explicit management of hidden states across rollouts and correct handling of \ac{BPTT}.

\section{Applications in Research}

\frameworkName was originally introduced to demonstrate the benefits of massively parallel simulation and GPU-based training for legged locomotion, achieving walking policies in only a few minutes~\citep{rudin2022learning}. Since then, it has become a foundation for a wide range of robotics research.

Building on this initial work, researchers have applied the framework for sim-to-real agile locomotion~\citep{cheng2023parkour, margolis2023walk} and whole-body control~\citep{fu2023deep, he2024hover,arm2024pedipulate}.
\citet{hoeller2024anymal} extended the framework to use mixed action distributions for training a high-level navigation policy that orchestrates multiple expert skills. \citet{rudin2025parkour} used teacher-student distillation to train a generalist locomotion policy. More recently, attention-based network architectures have been employed for locomotion on sparse terrain~\citep{he2025attention} and depth-based navigation~\citep{yang2025improving}.
By incorporating demonstrations, several works have extended the framework for adversarial style~\citep{vollenweider2022advanced} and DeepMimic style training~\citep{sleiman2024guided}. Other studies~\citep{lee2024learning, dadiotis2025dynamic} have enhanced the PPO implementation to handle constraints during training through P3O~\citep{zhang2022penalized} and CaT~\citep{chane2024cat}.
Finally, the symmetry-based augmentation methods included in the framework have been leveraged for multi-agent coordination~\citep{li2025marladona} and low-level control~\citep{zhang2024learning, zhang2024wococo}.

\section{When and When Not to Use \frameworkName}

\frameworkName is designed for robotics researchers who need a compact, modifiable, and well-validated learning codebase. It supports research aimed at advancing the state-of-the-art in robotics via \ac{RL}, allowing users to either implement new ideas or apply the library directly to enhance real-robot capabilities. Its features are specifically built to improve performance and facilitate successful sim-to-real transfer, a critical step for validating new methods.

The library is not intended for fundamental or general-purpose machine learning research. By design, it includes a limited set of algorithms, making it unsuitable for straightforward benchmarking of other \ac{RL} algorithms. Furthermore, \frameworkName does not provide native support for pure imitation learning.

%\newpage

% Acknowledgements and Disclosure of Funding should go at the end, before appendices and references
\acks{Nikita Rudin and David Hoeller developed the initial version (v1.0.2) of the library, as used in \citet{rudin2022learning}, during their time at ETH Z\"{u}rich and NVIDIA. Clemens Schwarke and Mayank Mittal have since maintained and extended the library. The authors thank all open-source contributors. This work is supported by NVIDIA.}

\bibliography{sample}

\end{document}

%% file: acronyms.tex
\acrodef{PPO}{Proximal Policy Optimization}
\acrodef{RL}{Reinforcement Learning}
\acrodef{BC}{Behavior Cloning}
\acrodef{RND}{Random Network Distillation}
\acrodef{BPTT}{Backpropagation Through Time}